\documentclass[a4paper,11pt,hidelinks]{article} 

\usepackage{hyperref}
\usepackage[left=1in, right=1in, top=1in, bottom=1in]{geometry}
\usepackage[utf8]{inputenc}
\usepackage[english]{babel}
\usepackage{multicol, multirow}
\usepackage{amsmath, amssymb}
\usepackage{url}
\usepackage{tabularx}
\usepackage{caption}
\usepackage{subcaption}
\usepackage[ruled,linesnumbered]{algorithm2e}
\usepackage{tablefootnote}
\usepackage{bm}
\usepackage{orcidlink}
\usepackage{natbib}
\usepackage{booktabs}
\usepackage{authblk}
\usepackage{float}

\usepackage{todonotes}

\setcitestyle{numbers,square}

\newcommand{\red}[1]{\textcolor{red}{#1}}
\newcommand{\blue}[1]{\textcolor{blue}{#1}}

\newcommand{\comment}[1]{}

\title{Robust personnel rostering: how accurate should absenteeism predictions be?}

\author[a,b,*]{Martina Doneda \orcidlink{0000-0002-1660-3100}}
\author[c]{Pieter Smet \orcidlink{0000-0002-3955-7725}}
\author[a]{Giuliana Carello \orcidlink{0000-0001-5163-9865}}
\author[d]{Ettore Lanzarone \orcidlink{0000-0001-8816-9086}}
\author[c]{Greet Vanden Berghe \orcidlink{0000-0002-0275-5568}}

\affil[a]{Politecnico di Milano, Department of Electronics, Information and Bioengineering, Milan, Italy}
\affil[b]{National Research Council, Institute for Applied Mathematics and Information Technologies, Milan, Italy}
\affil[c]{KU Leuven, Department of Computer Science, CODeS, Gent, Belgium}
\affil[d]{University of Bergamo, Department of Management, Information and Production Engineering, Dalmine, Italy}
\affil[*]{Corresponding author: martina.doneda@polimi.it}

\date{\today}

\begin{document}

\maketitle

\begin{abstract}

Disruptions to personnel rosters caused by absenteeism often necessitate last-minute adjustments to the employees' working hours.
A common strategy to mitigate the impact of such changes is to assign employees to reserve shifts: special on-call duties during which an employee can be called in to cover for an absent employee.
To maximize roster robustness, we assume a predict-then-optimize approach that uses absence predictions from a machine learning model to schedule an adequate number of reserve shifts. 
In this paper we propose a methodology to evaluate the robustness of rosters generated by the predict-then-optimize approach, assuming the machine learning model will make predictions at a predetermined prediction performance level.
Instead of training and testing machine learning models, our methodology simulates the predictions based on a characterization of model performance. 
We show how this methodology can be applied to identify the minimum performance level needed for the model to outperform simple non-data-driven robust rostering policies.
In a computational study on a nurse rostering problem, we demonstrate how the predict-then-optimize approach outperforms non-data-driven policies under reasonable performance requirements, particularly when employees possess interchangeable skills.

\end{abstract}

\section{Introduction}

Employee absenteeism is defined as the unplanned absence of an employee from work when they are scheduled to be present.
Statistics from 2022 report that the average short-term\footnote{Defined in the report as an absence lasting less than one month.} absenteeism rate in Belgium was $3.43\%$ \citep{SDWorx2022}.
The same study reports that 4.67\% of all working days in January of that same year were lost to short-term absences.
Employee absenteeism can be attributed to various interrelated factors such as health problems, challenges with work-life balance and workplace harassment \citep{tarro2020}.
Regardless of the root cause, absenteeism has important direct and indirect effects.
For example, reduced staffing levels are known to impact service quality and productivity \citep{hudson2015}.
Studies on the effects of absenteeism have shown that the negative impact is especially high when absent employees have specialized task-specific knowledge, when the work is highly interconnected (such as on assembly lines), or when companies are unable to substitute absent employees due to organizational limitations \cite{grinza2020impact}.

To repair disruptions in the employees' rosters caused by absenteeism, various \textit{rerostering} strategies have been proposed \citep{wickert2019}. 
Although computational experiments have demonstrated the positive organizational impact of these methods, repairing disruptions inevitably introduces personal discomfort \citep{kocakulah2016}.
Last-minute changes to an employee's roster may severely affect their personal life and negatively impact their job engagement and productivity \citep{ticharwa2019}.
Instead of reacting to disruptions, this paper focuses on proactively generating \textit{robust} personnel rosters that are immune to a certain level of employee absenteeism, thereby reducing the negative effect of last-minute changes needed to repair disruptions \citep{wickert2021}.
An intuitive way of generating robust rosters is to forecast employee absences and include this information in the roster construction process.
This predict-then-optimize approach begins by conceptualizing, training and testing a machine learning (ML) model to predict employee absenteeism. 
These predictions are subsequently included as parameters in an optimization model to generate personnel rosters. 
The quality of the solutions generated by the optimization model therefore depends on the predictive performance of the ML model \citep{elmachtoub2022}.
Intuitively, it is always better to have a more accurate prediction rather than a less accurate one. 
Yet, there is trade-off between model performance and model training costs \citep{tulabandhula2013} and, in some cases, performance itself has an upper bound. 
The trade-off is context-specific, and is often impractical to determine beforehand.
However, given the immense effort required to collect data and training ML models, it is worthwhile to estimate the potential benefit of such predictions in advance.

We investigate a methodology to determine the robustness of rosters generated using the aforementioned predict-then-optimize approach, assuming the employed ML model can make predictions at a predetermined performance level.
Rather than actually training and testing ML models, our proposed methodology involves \textit{simulating} the predictions an ML model would make at a given performance level.
By varying the prediction performance level, simulating the ML model's predictions and evaluating robustness of the resulting rosters, we can determine the prediction performance level needed to reach a given quality threshold.
As our methodology does not involve training and testing ML models, we do not require access to data to determine these minimum performance requirements.

\citet{farrington2023} introduced an interesting methodology for managing perishable inventory.
They simulated outcomes of a predictive ML model at various accuracy levels to determine when an ML model would lead to more efficient inventory management compared to a basic inventory allocation policy.
We adapt their approach of simulating predictions to the context of robust personnel rostering, highlighting the importance of this methodology in the predict-then-optimize paradigm.

\comment{
The methodology presented by \citet{farrington2023} is the most closely related to what we propose. 
In their work, the authors carry out several simulations to determine how good a predictive model must be to deploy a specific policy for the management of the blood bank in a British hospital. 
The main focus is on understanding which policy is more effective in function of the (partial) knowledge they may have of future events.
\red{We take this approach as inspiration and present a more general methodology capable of simulating ML predictions at specified performance levels, which is then applied to the robust personnel rostering problem. }
\blue{With respect to \cite{farrington2023}, in this work we present two novel contributions: \textit{i}) the focus on the importance simulated machine learning can have from a methodological standpoint, and \textit{ii}) its application to robust scheduling.}
}

The remainder of this paper is organized as follows. 
Section \ref{s:lit} reviews the literature related to robustness in personnel rostering. 
Section \ref{s:prob} describes the considered rostering problem and introduces how robustness can be included in rosters. 
Section \ref{s:framework} introduces the methodology to simulate predictions and applies it to robust rostering.
Sections \ref{s:plan} and \ref{sec:results} provide information concerning the computational study and a discussion of the results, respectively.
Finally, Section \ref{s:conc} concludes the paper and identifies promising directions for future research.

\section{Related work on robust rostering}
\label{s:lit}

There are two general strategies in the literature for generating rosters that are robust with respect to employee absenteeism. 
Horizontal strategies introduce robustness by allowing overtime or by enabling shifts to partially overlap in order to ensure a degree of redundancy \citep{ingels2018}. 
On the other hand, vertical strategies enforce robustness through resource buffers that store additional employees. 
Generally, two types of buffer can be distinguished.
Capacity buffers are created by assigning more employees than necessary to cover a nominal demand \citep{ingels2019}.
The second type of buffer is created by assigning employees to reserve shifts \citep{ingels2015}.
These are special non-working shifts during which employees are on-call and can be relied upon to work a regular shift if the need arises.
Clearly, this type of buffer is much more flexible than a capacity buffer, as reserve shifts can be converted into any other shift \citep{potthoff2010column}.
Moreover, they are usually less expensive for organizations, as the compensation employees receive for being on-call is typically far less than working a regular shift or overtime.
However, while it may be less costly for the organization, the unpredictability of being called in to work has been shown to have negative consequences, regardless of whether or not employees are actually called in \citep{bamberg2012effects}.

Various optimization models exist to optimize the allocation of reserve shifts to employees.
\citet{ingels2015} conduct a computational study to investigate how reserve shifts affect roster robustness when both demand and capacity are uncertain.
They analyze five strategies for scheduling reserve shifts using a combination of specific demand requirements and time-related constraints.
They evaluate the robustness of the resulting rosters by means of a discrete-event simulation.
Their experiments identify a trade-off between the wage costs associated with scheduling reserve shifts and staff shortages.

\citet{dillon1999us} investigate the use of reserve shifts in airline crew scheduling.
Reserve pilots and flight attendants are on standby to substitute for crews who cannot operate their assigned flights due to either illness or the delay/cancellation of connecting flights.
An automated decision support system is presented to outsource the work to external personnel whose wages are many times higher than those of regular employees.

\citet{potthoff2010column} developed a column generation algorithm to reassign Dutch railway personnel to cope with large disruptions. Their algorithm reassigns duties and defines any additional task needed (i.e. \textit{deadheading}) to minimize the cost to restore the functioning of the network. 

\citet{becker2019cyclic} propose an algorithm to generate cyclic rosters with reserve shifts.
To ensure fairness, the reserve shift assignments rotate after each cycle of the regular shifts.
The algorithm was applied to a German emergency medical services provider and was shown to be able to take into account employee preferences related to weekend work, recovery times, and fairness.

\citet{el2016proactive} investigate how scheduling reserve shifts can address issues related to overcrowding in emergency departments.
They propose a scheduling policy that balances demand coverage and labor cost.
The problem is modeled as a two-stage stochastic programming problem where the first stage rosters employees based on estimations of demand, while the second stage involves the day-to-day decisions.
In a series of computational experiments, they analyze the advantages and disadvantages of reserve shifts for emergency departments under different demand scenarios.

\citet{wickert2021} introduce two metrics of roster robustness based on the characteristics of the roster itself.
By including these metrics in an optimization model, they generated rosters of varying degrees of robustness.
A computational study demonstrates how reserve shift buffers are generally preferred over capacity buffers because they are less expensive and more flexible when used to repair disruptions.

\section{Personnel rostering problem description}
\label{s:prob}

Several weeks before a scheduling period begins, the rostering problem is solved so that employees are aware of their working hours well in advance.
A new roster is generated by assigning shifts to employees.
The rostering problem we consider is based on the general problem definition proposed by \citet{ceschia2019}.
Let $N$ denote the set of employees, $K$ the set of employee skills, $S$ the set of shifts and $D$ the set of days in the scheduling period.
For each day $d \in D$, shift $s \in S$ and skill $k \in K$, the minimum number of employees that is required to be present $m_{dsk}$ is given.
This demand is treated as a soft constraint: if it cannot be met, we assume it is possible to call in external personnel whose wages are many times higher than the wages of regular employees. 

An employee can be assigned at most one shift per day.
Each employee $n \in N$ is qualified for a subset of skills $K_n \subseteq K$.
A shift that requires a specific skill can only be assigned to an employee who is qualified for that skill.
To ensure sufficient resting time between two working days, the set $\tilde{S}$ contains pairs of shifts $(s_1,s_2)$ that cannot be assigned to the same employee on two consecutive days.
The remaining constraints are related to the employees' contracts and personal preferences.
Each employee $n \in N$ has to work between $\beta^3_n$ and $\beta^5_n$ days in the scheduling period.
Overtime incurred by assigning more shifts than the maximum number $\beta^5_n$ is allowed.
However, undertime (assigning fewer than $\beta^3_n$ shifts) is forbidden.
The number of consecutive working days must not exceed $\beta^1_n$, while the number of consecutive nights shifts is limited to $\beta^2_n$.
Finally, a set $U$ of tuples $(n,d,s)$ specifies that employee $n$ has requested not to work shift $s$ on day $d$.
The assignments made in the previous scheduling period are taken into account to prevent violations of constraints concerning consecutive assignments at the beginning of the current period.

The objective function is a weighted sum of the employees' wage costs (including overtime) and the wages of external employees needed to cover understaffing.
The wage cost of employee $n$ for working a single day is denoted by $\omega^1_n$.
For each day worked in excess of the maximum number $\beta^5_n$, an overtime cost $\omega^5_n$ is incurred.
The daily wage cost of an external employee is denoted by $\omega^6$.

To ensure an unambiguous understanding of the problem, Appendix \ref{ss:ro-mip} provides a formal definition of it as a mixed integer programming (MIP) formulation.

\subsection{Robust rostering}
\label{ss:robust_rostering}

When an employee becomes absent during the scheduling period, a rerostering problem must be solved.
In contrast to the rostering problem, which involves generating a new roster from scratch, the rerostering problem modifies an existing roster to repair disruptions caused by absences.
To reduce the impact of such disruptions, we generate robust rosters by using reserve shift buffers.
Appendix \ref{ss:re-mip} provides details concerning the specific rerostering problem we consider in this work and describes how we employ the reserve shift buffers when rerostering.

Let $c^*_d$ be the number of reserve shifts that must be included in the roster on day $d$.
The required number of reserve shifts per day is treated as a soft constraint, whose violation is penalized in the objective function with a penalty $\omega^r$.
The wage cost associated with assigning a reserve shift to employee $n$ is denoted by $\omega^7_n$.
Each employee $n \in N$ can be assigned to at most $\beta^6_n$ reserve shifts during the scheduling period.
The employees' contractual constraints are defined in such a way that generated rosters remain feasible for any possible conversion of the reserve shifts.
Regardless of whether the reserve shift is converted into a working shift during rerostering or whether it ultimately remains unused, no contractual constraint will be violated.

Formulated as a MIP problem, the objective function of the rostering MIP formulation \eqref{eq:OF_ro}-\eqref{eq:ro_bounds3} is replaced with Equation \eqref{eq:obj_complete}.
Let $s^{\prime} \in S$ be the index of the reserve shift in $S$.
The binary decision variable $x_{ndsk}$ equals one if employee $n$ is assigned to shift $s$ on day $d$ using skill $k$, and zero otherwise.
The non-negative, continuous variable $v^r_d$ counts the shortfall in assigned reserve shifts on day $d$.
\begin{equation}
 \min \eqref{eq:OF_ro} + \sum_{n \in N}\sum_{n \in D}\sum_{k \in K} x_{nds'k}\omega^7_n + \sum_{d \in D}v^r_d\omega^r
 \label{eq:obj_complete}
\end{equation}

Additionally, two new constraints are included in the MIP formulation.
Constraints \eqref{eq:max_reserve_shifts} limit the maximum number of assignments to the reserve shift for each employee, while Constraints \eqref{eq:reserve_robustness} ensure that at least $c^*_d$ reserve shifts are included in the roster on day $d$.
\begin{align}
 & \sum_{d \in D}\sum_{k \in K}x_{nds'k}\leq \beta^6_n & \forall n \in N \label{eq:max_reserve_shifts} \\
 & \sum_{n \in N}\sum_{k \in K}x_{nds'k} + v^r_d \geq c^*_d & \forall d \in D \label{eq:reserve_robustness}
\end{align}

The number of reserve shifts required on each day can be determined in various ways.
For example, by consulting a human expert or by analyzing the outcome of a series of simulations \citep{wickert2021}.
A third approach involves using an ML model to make predictions, which then feature as parameters in the optimization model.
This \textit{predict-then-optimize} approach to robust rostering begins by conceptualizing, training and testing an ML model to predict employee absences.
Based on these predictions, the number of reserve shifts on each day is determined.
The resulting $c^*_d$ parameter values are then included the rostering model.

A critical characteristic of the predict-then-optimize approach is that the generated roster's robustness may depend heavily on the ML model's prediction performance. 
The highest reachable performance level of an ML model may be limited by the availability of data or the costs related to data collection and model training.
However, this performance level may be insufficient to benefit the system.
Conversely, the ML model may have been needlessly over-trained to attain extremely high performance levels, resulting in excessive model training costs.

In the following section we propose a methodology to compute the robustness of a roster generated by a predict-then-optimize approach, assuming the ML model can make predictions at a predetermined performance level.
We then show how we can employ this methodology to determine the minimum performance requirements needed to obtain sufficiently robust rosters.
The key advantage of our methodology is that it does not involve training ML models or require extensive data on the phenomenon.
Instead, we propose a way of simulating the predictions a model would make at a given prediction performance level.
We call this methodology \textit{simulated} ML in order to distinguish it from traditional ML.

\section{Simulated ML for robust rostering}
\label{s:framework}

The ML model in the predict-then-optimize approach described in Section \ref{ss:robust_rostering} is a binary classifier: its output is either equal to 1 or to 0, predicting whether or not an absence will occur for each day and employee.
A \textit{confusion matrix} \citep{kuhn2022} enables a comparison of observed and predicted values for binary classification problems, as shown in Figure \ref{tab:confusion}.
We use the following two metrics derived from the confusion matrix to characterize prediction performance of the binary classifier:

\begin{itemize}
 \item Sensitivity ($\alpha$), or True Positive Rate (TPR): the probability of a positive classification in positive observations. Calculated as $TP/(TP+FN).$
 \item Specificity ($\beta$), or True Negative Rate (TNR): the probability of a negative classification in negative observations. Calculated as $TN/(TN+FP).$ $1-\beta$ is called False Positive Rate (FPR).
\end{itemize}

\begin{figure}[hptb]
 \centering
 \footnotesize
 \bgroup
 \def\arraystretch{1.5}
 \begin{tabular}{l|l|c|c|}
 \multicolumn{2}{c}{} & \multicolumn{2}{c}{\textbf{Observed}} \\
 \cline{3-4}
 \multicolumn{2}{c|}{} & \textbf{Positive} & \textbf{Negative}\\
 \cline{2-4}
 \multirow{2}{*}{\textbf{Predicted}}& \textbf{Positive} & True positive (TP) & False positive (FP)\\
 \cline{2-4}
 & \textbf{Negative} & False negative (FN) & True negative (TN) \\
 \cline{2-4}
 \end{tabular}
 \egroup
 \caption{Confusion matrix standard notation.} 
 \label{tab:confusion}
\end{figure}

In the context of predicting absences, we are interested in the prediction of a binary event with strong class imbalance, meaning that the frequency of the event of interest is relatively low.
A priori, we do not know whether a single prediction made by the ML model is right or wrong.
We only know the frequency $\rho$ with which the uncertain rare event occurs.
Hence, we expect the frequency of predictions of the positive class to be around $\rho$, regardless of how poor the performance of the ML model is.
Therefore, we scale the FPR by the event frequency $\rho$, referred to as the rescaled FPR (rFPR).

\begin{figure}[hptb]
 \centering
 \includegraphics[width=0.9\textwidth]{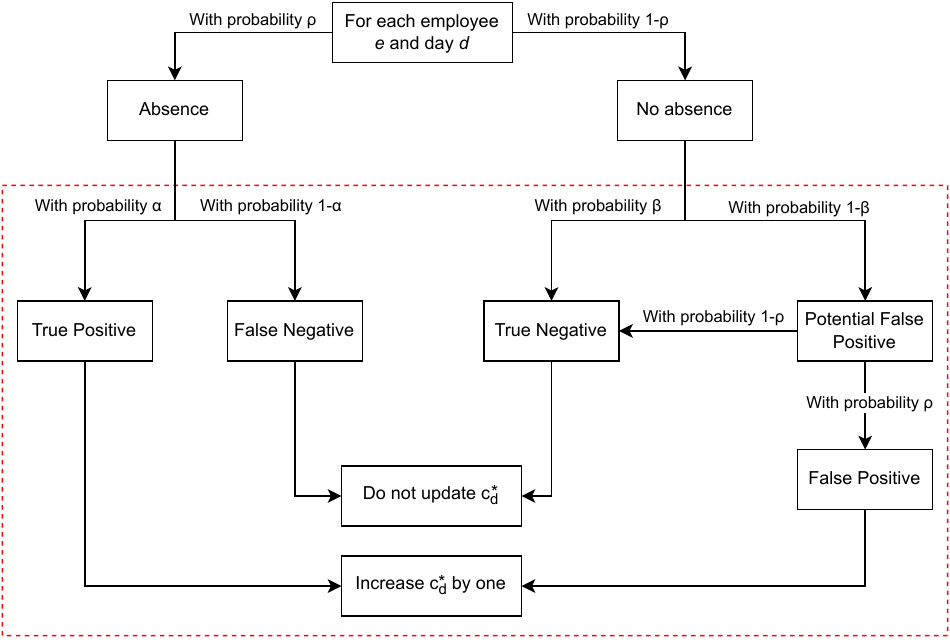}
 \caption{Schematic overview of how a simulated binary classifier determines the number of reserve shifts $c^*_d$ for each day $d \in D$ for a given $\alpha$ and $\beta$. Steps located within the red dashed rectangle form the fundamental components of the simulated ML methodology.}
 \label{img:ml_overview}
\end{figure}

By considering different values for $\alpha$ and $\beta$, it is possible to characterize binary classifiers with different performances.
Figure \ref{img:ml_overview} provides a schematic overview of how a simulated binary classifier determines the integer value of $c^*_d$ in Constraints \eqref{eq:reserve_robustness} for a given $\alpha$ and $\beta$.
For each employee $e$ on each day $d$, we simulate their absence with probability $\rho$. 
An absence will be correctly predicted by the classifier with probability $\alpha$, resulting in an increase of the number of required reserve shifts $c^*_d$ by one.
With probability $1-\alpha$, the model will incorrectly predict an absence, resulting in a False Negative.
In this case, no action is taken and the value of $c^*_d$ is unaffected.
Similarly, with probability $\beta$, the model will correctly predict the negative class, resulting in a True Negative and no additional reserve shifts.
With probability $1-\beta$, the model incorrectly predicts the negative class, resulting in a so-called \textit{potential} False Positive.
Given that we know an absence will occur with probability $\rho$, we use this value to determine whether a potential False Positive becomes an actual False Positive or not.
This way, the overall number of predicted absences will be reasonable even when $\beta$ assumes very small values. 

After completing these steps, a robust roster can be generated by solving the MIP problem provided in Appendix \ref{ss:re-mip} using the obtained $c^*_d$ values.
Note that the identities of the employees that are absent are not known by the optimization model. 
The only parameter that is passed to the model is the number $c^*_d$ of predicted absences for a given day $d$.
To compute the rerostering cost for the roster obtained for given $\alpha$ and $\beta$ values, we include the real absences in the generated roster and solve the resulting rerostering problem.
By comparing this value against a predefined threshold cost, we can adjust $\alpha$ and $\beta$ and repeat the process until we obtain a satisfactory rerostering cost.


\section{Computational study}
\label{s:plan}

This section introduces the experimental setup used in the computation study on the conditions in which ML predictions can improve robust rostering for a problem involving rostering nurses in a hospital ward.
Section \ref{sec:data} describes the data used in the computational study.
Section \ref{sec:evaluation} introduces the evaluation metrics that were used.

\subsection{Data} \label{sec:data}

We conduct our computational experiments using the problem instances introduced by \citet{wickert2021}.
These two instances were derived from the second International Nurse Rostering Competition \citep{ceschia2019}, thus including a set of constraints and problem characteristics that are often encountered in practice.
The problem instances consist of 35 nurses and a planning horizon of four weeks.
There are four shift types (early, late, day and night) in addition to the reserve shift.
Table \ref{tab:values} provides the values of the different weights of the robust rostering and rerostering objective functions used in the experiments.
Unexpected last-minute calls to nurses with a day off or changing their assigned working shift typically have a strong negative impact on their personal lives.
These weights therefore reflect the preference of converting a reserve shift into a working shift over converting a day off into a working shift or changing the shift of an already scheduled nurse.

The first problem instance considers employees with uniform skills.
This instance considers only a single skill and all employees are qualified for this skill.
By contrast, the second instance has different employee types representing a hierarchical skill structure: head nurse, nurse, trainee and caretaker.
These types are organized in such a way that substitutions based on their skills can occur: head nurses can substitute for nurses and caretakers, while nurses can substitute for caretakers. 
Caretakers and trainees cannot substitute for any other employee type.
When there are no skills, the degree of substitutability between employees is maximized: any employee can substitute for any other.
However, when considering hierarchical skills, substitutability is decreased and the rerostering process typically has less decision flexibility.

\begin{table}[hptb]
 \footnotesize
 \centering
 \begin{tabular}{cl}
 \toprule
 \multicolumn{2}{l}{\textbf{Robust rostering costs and weights}} \\
 \midrule
 $\omega^1_n$ & $\{100, 70, 50, 30\}$ \tablefootnote{For head nurses, nurses, caretakers and trainees, respectively.} \\
 $\omega^5_n$ & $1.5\cdot\omega^1_n$ \\
 $\omega^6$ & $5\cdot \max{\omega^1_n}$ \\
 $\omega^7_n$ & $0.1\cdot\omega^1_n$\\
 $\omega^r$ & $1\text{e+}3$ \\
 \midrule
 \multicolumn{2}{l}{\textbf{Rerostering costs and weights}} \\
 \midrule
 $\omega^2_n$ & $\omega^1_n$ \\
 $\omega^3_n$ & $0.1\cdot\omega^1_n$\\
 $\omega^4_n$ & $1.5\cdot\omega^1_n$ \\
 \bottomrule
 \end{tabular}
 \caption{Robust rostering and rerostering costs used in the computational study.}
 \label{tab:values}
\end{table}

To evaluate the binary classifier at various performance levels, we consider the following values for TPR and rFPR: 0.0, 0.1, 0.2, 0.3, 0.4, 0.5, 0.6, 0.7, 0.8, 0.9, 1.0. 
A prediction model with $\text{TPR} = 0$ and $\text{rFPR} = 0$ is equivalent to never enforcing any reserve shift, because no True or False Positives are ever predicted.
Conversely, if $\text{TPR} = 1$ and $\text{rFPR} = 0$ then the model is always able to correctly classify each absence or non-absence.

\subsection{Evaluation}\label{sec:evaluation}

Two cost metrics are employed to evaluate the rosters generated in the computational study.
The \textit{rostering cost} refers to the value of the objective function of the robust rostering model detailed in Equation \eqref{eq:obj_complete}.
Meanwhile, the \textit{rerostering cost} refers to the objective function value of the rerostering model detailed in Equation \eqref{eq:OF_re}.
Note that this consists of the rostering costs plus the costs incurred by any changes made to that roster.
To compute the rerostering cost, the problem described in Appendix \ref{ss:re-mip} was solved.
Employee absences were generated using a Bernoulli distribution with $p = \rho$ for each nurse.
Similar to \citet{wickert2021}, we use an average absenteeism rate of $\rho = 2.64\%$.
In total, 100 absenteeism scenarios for each problem instance were generated.
The reported rerostering costs are the averages over these 100 scenarios.

The threshold value used to determine whether or not the rerostering cost obtained is satisfactory is computed using a non-data-driven robust rostering policy that is defined based on the results of \citet{wickert2021}.
This \textit{baseline} policy does not make use of predictions on the values of uncertain parameters, and instead assigns one, two, three or four reserve shifts on each day of the scheduling period.

We ran all experiments on an AMD Ryzen 9 5950X 16-core processor at 3.40 GHz with 64 GB of RAM. 
The integer programming problems were solved using Gurobi 10.0.3 with the default optimality gap $1\text{e-}4$ and a maximum computation time of $100$ seconds.

\section{Results}\label{sec:results}

All rostering and rerostering problems with uniform skills were solved to optimality. 
The average computation time required for solving one rostering problem instance was $0.20$ seconds, while the average computation time for solving one rerostering problem was $0.27$ seconds.
For the problem instances with hierarchical skills, all but one were solved to optimality within the time limit.
The one non-optimal solution had an optimality gap of $1.35\text{e-}3\%$. 
Excluding this single non-optimal instance, the average computation time for solving the rostering problem was $0.5$ seconds, while the average computation time for rerostering was $2.35$ seconds.


\subsection{Uniform skills}
\label{ss:results1}

\begin{figure}[hptb]
\centering
\begin{subfigure}{0.45\textwidth}
 \includegraphics[width=1\textwidth]{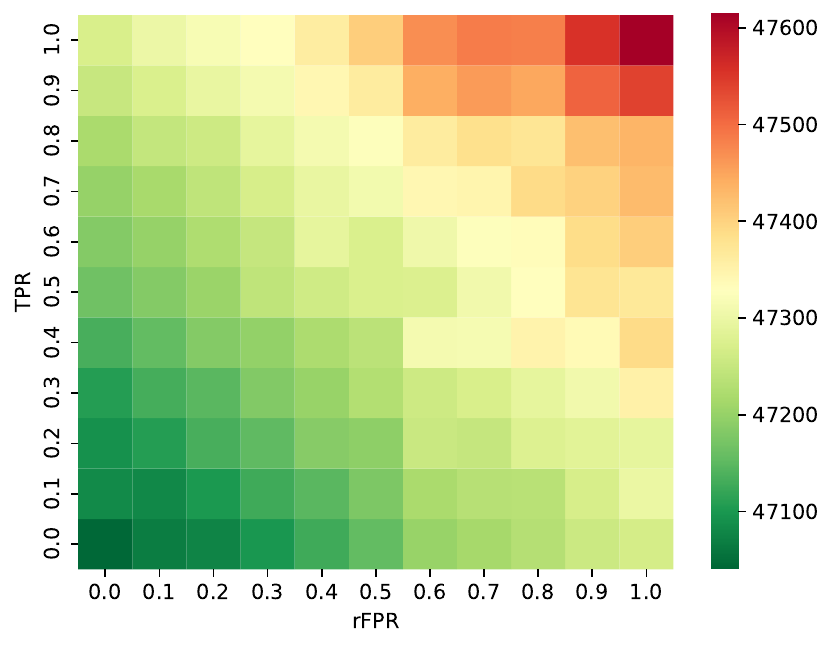}
 \caption{Rostering cost}
 \label{img:single-ro}
\end{subfigure}
\begin{subfigure}{0.44\textwidth}
 \includegraphics[width=1\textwidth]{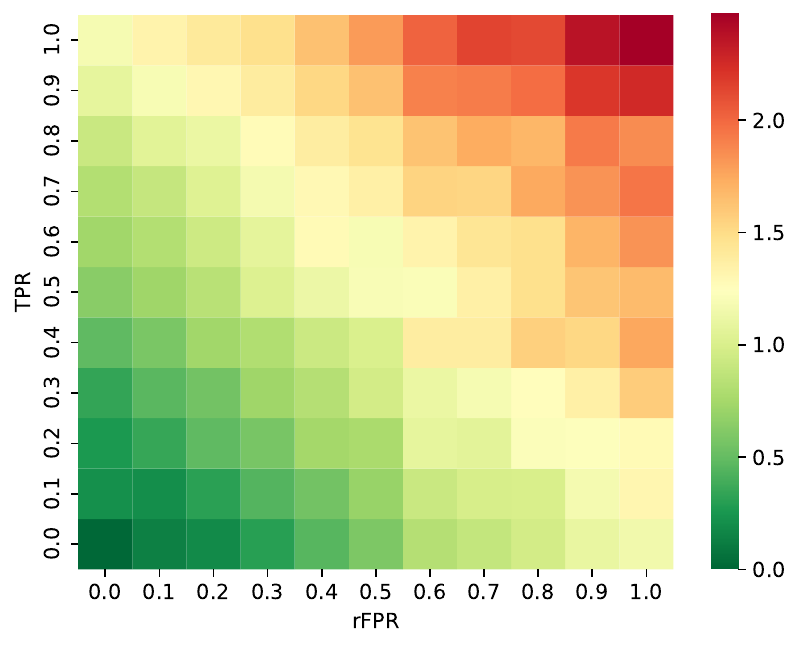}
 \caption{Number of reserve shifts per day}
 \label{img:single-number-res}
\end{subfigure}
\caption{Rostering costs and number of scheduled reserve shifts at various performance levels of the prediction model for the problem instance with uniform skills.}
\label{img:rore}
\end{figure}

Figure \ref{img:single-ro} plots the obtained rostering costs for different values of the TPR and rFPR.
The lowest rostering costs are observed when the TPR and rFPR are both small.
Under these conditions, the ML model predicts few absences and thus the solution includes few reserve shifts, resulting in an overall low rostering cost.
This trend is also evident in Figure \ref{img:single-number-res}, which shows the average number of scheduled reserve shifts on each day.
When the TPR or rFPR increases, the rostering cost also increases as more (true or false) absences are predicted and therefore more reserve shifts are included in the roster.
The highest rostering costs are obtained when both the TPR and rFPR are large.

\begin{figure}[hptb]
\centering
\begin{subfigure}{0.46\textwidth}
 \includegraphics[width=1\textwidth]{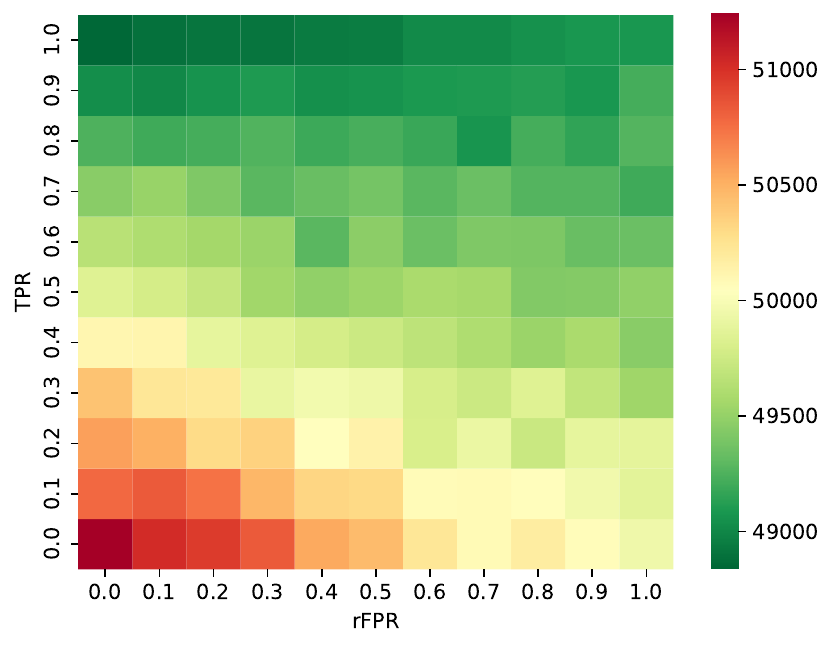}
 \caption{Rerostering cost}
 \label{img:single-re}
\end{subfigure}
\begin{subfigure}{0.44\textwidth}
 \includegraphics[width=1\textwidth]{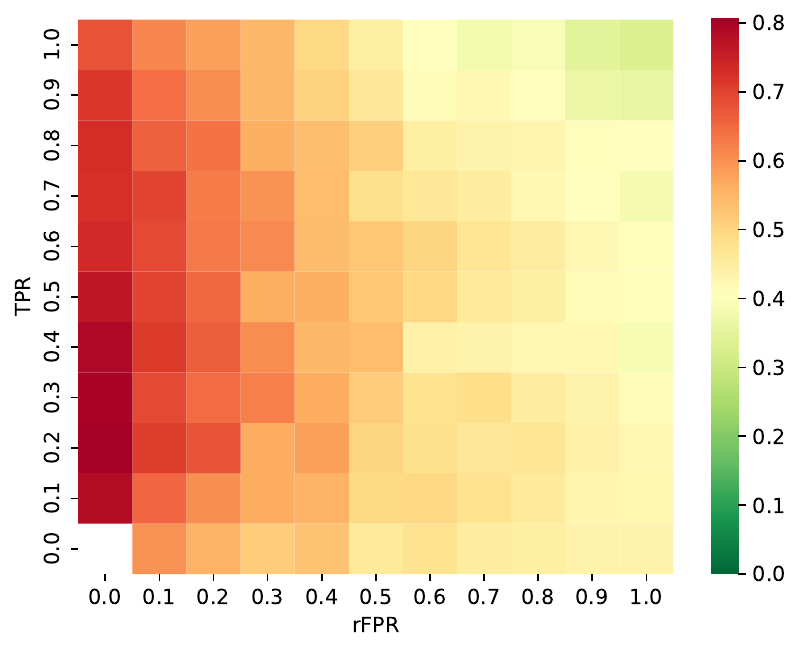}
 \caption{\% of reserve shifts changed}
 \label{img:single-res}
\end{subfigure}
\begin{subfigure}{0.44\textwidth}
 \includegraphics[width=1\textwidth]{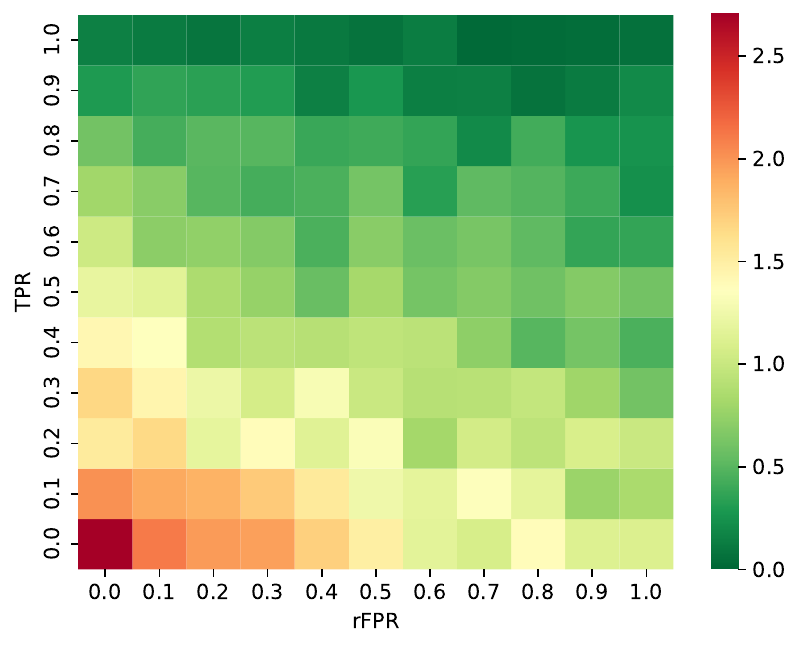}
 \caption{Number of working shifts changed}
 \label{img:single-work}
\end{subfigure}
\begin{subfigure}{0.44\textwidth}
 \includegraphics[width=1\textwidth]{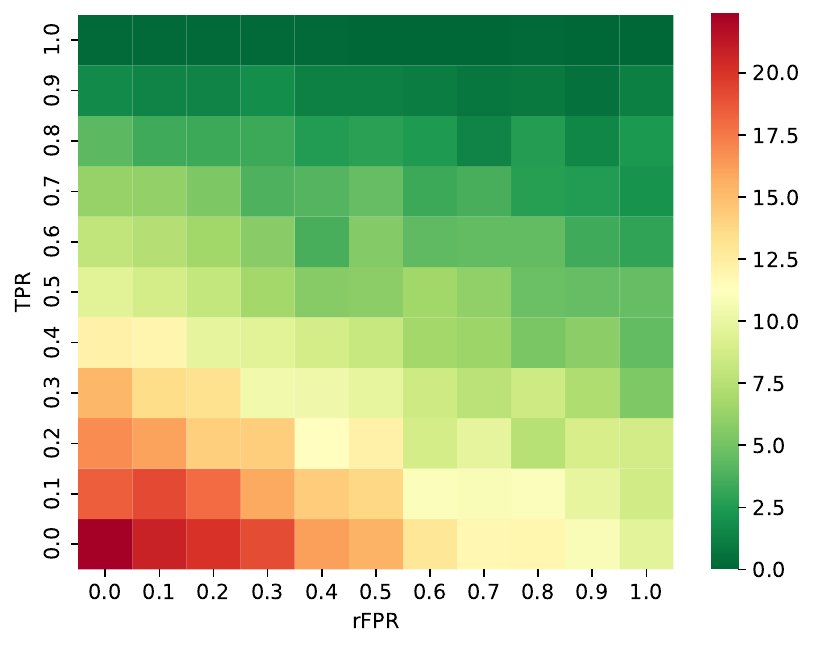}
 \caption{Number of days off changed}
 \label{img:single-off}
\end{subfigure}
\caption{Rerostering costs and the number of changes made during rerostering at various performance levels of the prediction model for the problem instance with uniform skills.}
\label{img:single-changes}
\end{figure}

Figure \ref{img:single-changes} provides insights into how the additional costs incurred during rerostering are affected by the TPR and rFPR.
Figure \ref{img:single-re} shows the rerostering cost for different values of TPR and rFPR, while Figures \ref{img:single-res}, \ref{img:single-work} and \ref{img:single-off} show how many reserve shifts, working shifts and days off were changed during rerostering, respectively.
The highest rerostering costs are obtained when both the TPR and rFPR are low.
Under these conditions, many False Negatives and True Negatives are predicted, resulting in few reserve shifts in the roster.
Consequently, in order to repair the roster, the rerostering method has to resort to changing working shifts or days off, as confirmed by the results in Figures \ref{img:single-work} and \ref{img:single-off}.
As the TPR increases, more True Positives are predicted and the rerostering cost decreases.
However, the rerostering cost also decreases for increasing values of rFPR.
Even for low TPR values, low rerostering costs are observed when the rFPR is high.
This demonstrates how the rerostering model can make good use of the reserve shifts to cover for absences, even if they were not initially assigned based on correct predictions.

Figure \ref{img:single-res} shows how the number of changes to reserve shifts is primarily driven by the rFPR: as the rFPR increases, fewer reserve shifts are converted.
The TPR has no identifiable impact on the number of reserve shifts converted during rerostering.
However, Figures \ref{img:single-work} and \ref{img:single-off} do clearly demonstrate how the number of changes to working shifts and days off are affected by both the rFPR and TPR.
The more reserve shifts that are included in a roster, the fewer working shifts and days off must be changed.
Even without properly positioning the reserve shifts on the day where absences will ultimately occur, the rerostering model can still benefit from them to avoid making other changes.

The observed decrease in rerostering cost for larger values of rFPR, in particular when the value of the TPR is low ($\alpha \leq 0.3)$, can be explained as follows.
Given the cost structure in these experiments, scheduling a reserve shift is relatively inexpensive.
However, given that there are no skills that limit the substitutions that can take place, reserve shifts are capable of covering any nurse absence for a certain shift.
This implies that, in some cases, a theoretically improperly-placed reserve shift can be used to cover for absent nurses that were not correctly identified by the ML model.
For example, assume the classifier was unable to correctly predict the absence of nurse $n$ who was assigned to shift $s$ of day $d$.
At the same time, the ML model had mistakenly predicted the absence of nurse $n^{\prime}$. 
The reserve shift originally planned to cover for nurse $n$ can still be used to cover for nurse $n^{\prime}$ at no additional cost.

\begin{figure}[hptb]
\centering
\begin{subfigure}{0.45\textwidth}
 \includegraphics[width=1\textwidth]{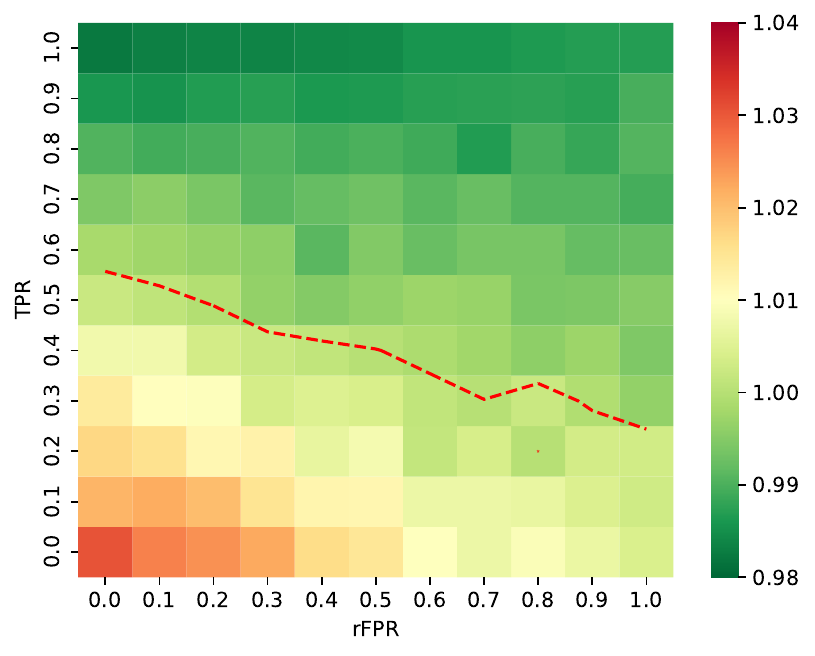}
 \caption{One reserve shift per day}
 \label{img:comp1}
\end{subfigure}
\begin{subfigure}{0.45\textwidth}
 \includegraphics[width=1\textwidth]{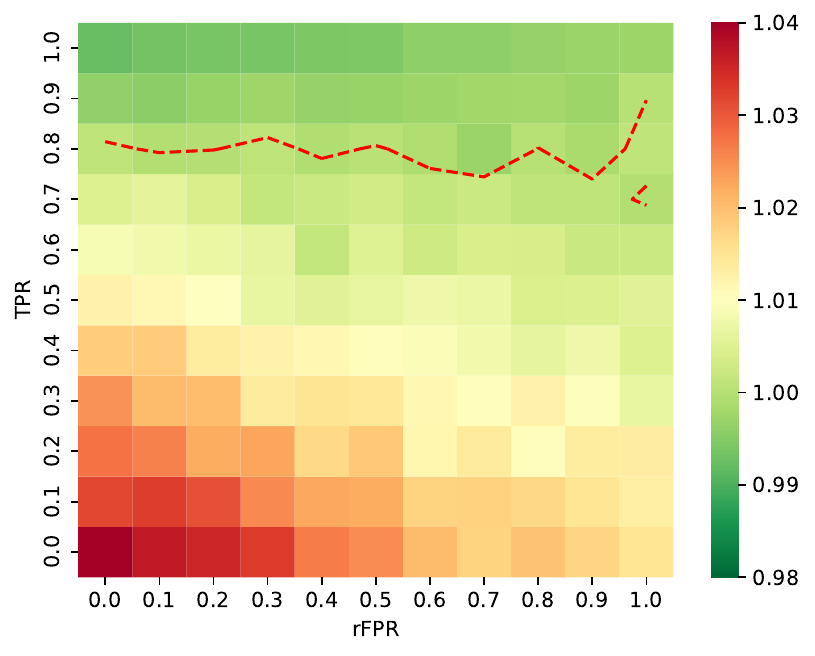}
 \caption{Two reserve shifts per day}
 \label{img:comp2}
\end{subfigure}
\begin{subfigure}{0.45\textwidth}
 \includegraphics[width=1\textwidth]{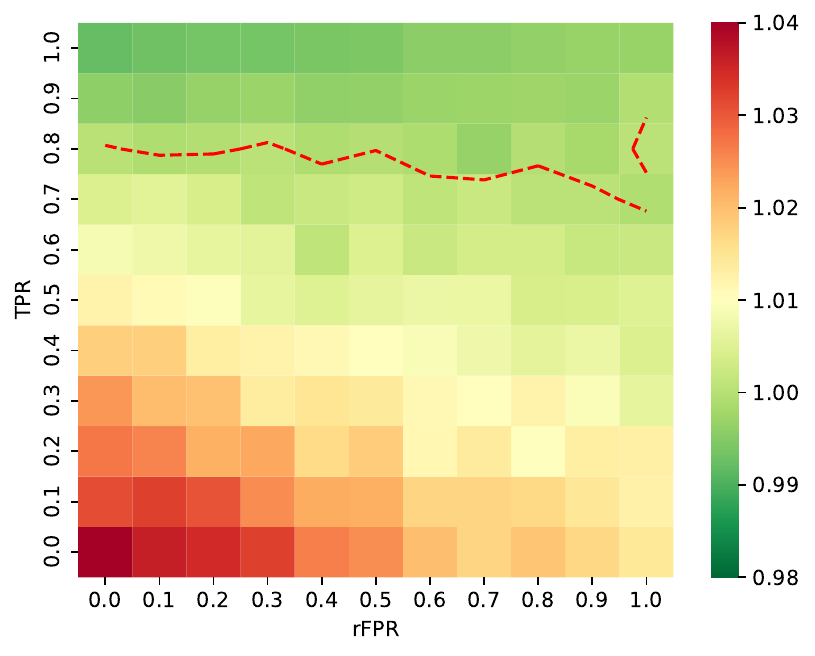}
 \caption{Three reserve shifts per day}
 \label{img:comp3}
\end{subfigure}
\begin{subfigure}{0.45\textwidth}
 \includegraphics[width=1\textwidth]{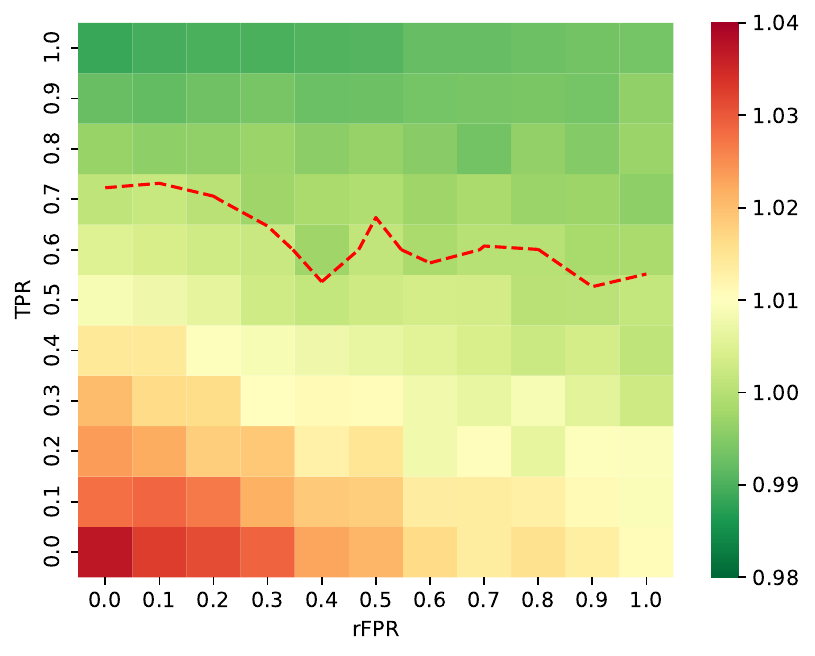}
 \caption{Four reserve shifts per day}
 \label{img:comp4}
\end{subfigure}
\caption{Rerostering cost of the ML-informed approach compared to the rerostering cost of the baseline policies for the problem instance with uniform skills.}
\label{img:single-comp}
\end{figure}

Figure \ref{img:single-comp} compares the ML-informed robust rostering approach against the four aforementioned baseline reserve shift scheduling policies that involve scheduling 1, 2, 3 and 4 reserve shifts per day.
The reported values are the ratio of the rerostering cost obtained by the ML-informed approach over the rerostering cost obtained by one of the baseline policies.
In these plots, a value equal to one represents conditions under which both the ML-informed approach and the baseline policy result in equal rerostering costs.
To better illustrate the performance comparison between the different approaches, a red dashed line also indicates when the ratio is equal to one. 
If the ratio is greater than one then the baseline approach generates less costly solutions, and vice versa.

When comparing against the policy that assigns one nurse to a reserve shift per day (Figure \ref{img:comp1}), we observe that even for relatively small TPR values, the ML-informed approach results in lower rerostering costs.
When two or three reserve shifts are assigned each day (Figures \ref{img:comp2} and \ref{img:comp3}), higher values of TPR are required for the ML model to outperform the baseline policies.
However, when four nurses are assigned on each day, the TPR requirement again becomes slightly less strong. The reason for this is that the four-nurse baseline policy already has a very high rostering cost to begin with, given that it schedules more reserve shifts than required during rerostering. The assumed cost for reserve shifts is relatively low, but still not negligible.

\subsection{Hierarchical skills}
\label{ss:results2}

While there is typically a lot of flexibility during rerostering when there are no skills to consider, rerostering with hierarchical skills is generally much more constrained.
The results discussed in Section \ref{ss:results1} demonstrate how even when absences are predicted to occur for the wrong nurse or day, the scheduled reserve shifts can still be beneficial when rerostering.
However, this does not hold in a setting with hierarchical skills, as nurses are no longer identical and always substitutable.

\begin{figure}[hptb]
\centering
\begin{subfigure}{0.45\textwidth}
 \includegraphics[width=1\textwidth]{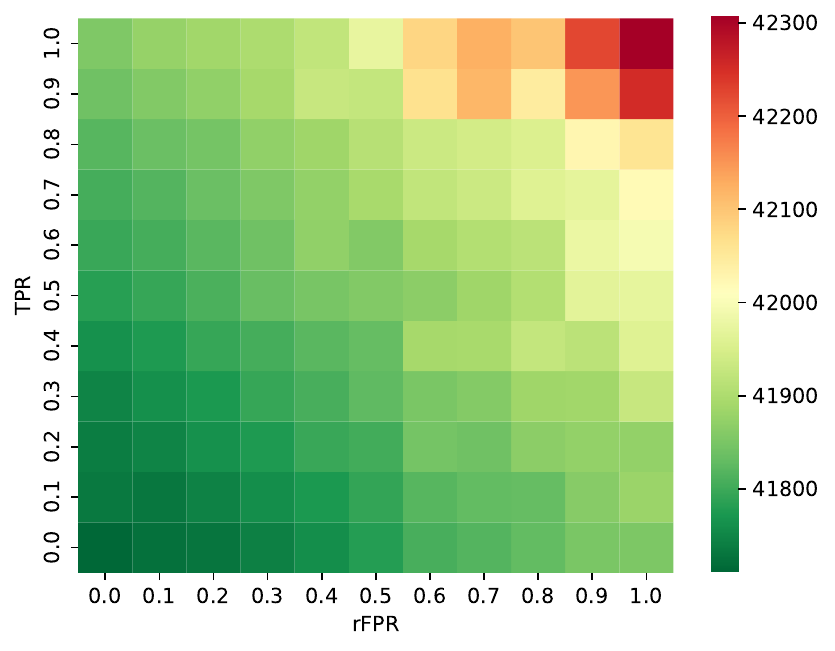}
 \caption{Rostering cost}
 \label{img:multi-ro}
\end{subfigure}
\begin{subfigure}{0.44\textwidth}
 \includegraphics[width=1\textwidth]{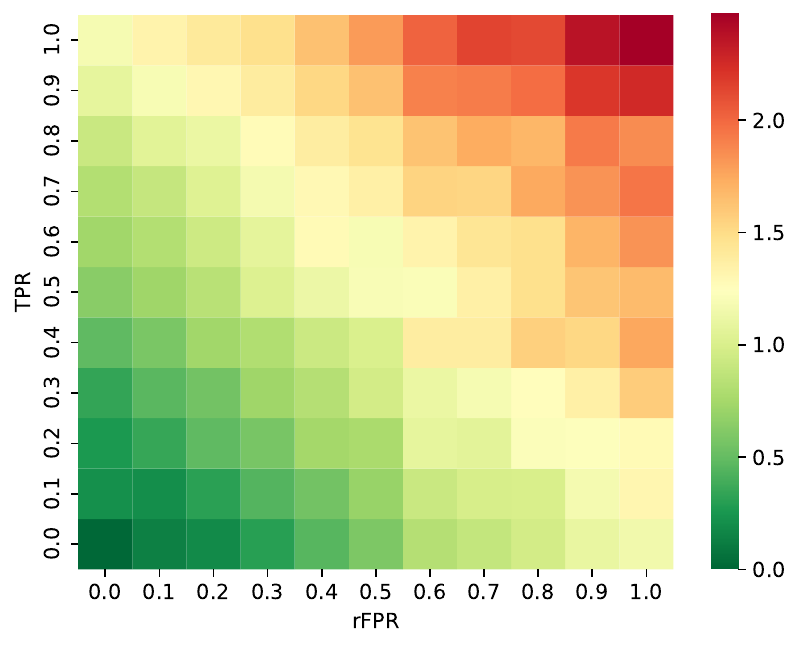}
 \caption{Number of reserve shifts per day}
 \label{img:multi-number-res}
\end{subfigure}
\caption{Rostering costs and number of scheduled reserve shifts at various prediction model performance levels for the problem instance with hierarchical skills.}
\label{img:multi-rore}
\end{figure}

Figure \ref{img:multi-rore} shows how the rostering cost and the number of scheduled reserve shifts changes for different values of the TPR and rFPR.
Similar to the scenario with uniform skills, more reserve shifts are scheduled when more absences are predicted (when the TPR or rFPR increase).
Due to the way the objective function is defined, reserve shifts will be assigned to the least costly nurses (caretaker of trainee) whenever possible.
However, the least costly nurses are also those who are unable to cover for more qualified personnel. 
The sudden increase in rostering cost, which can be seen in Figure \ref{img:multi-ro}, occurs when the least expensive nurses have all been assigned to reserve shifts and the rostering model is forced to assign more qualified (and thus costly) nurses.
The number of scheduled reserve shifts on each day is comparable to the setting with uniform skills (Figure \ref{img:multi-number-res}), given that this value does not depend on the nurses' skills but only on the TPR and rFPR.

\begin{figure}[hptb]
\centering
\begin{subfigure}{0.45\textwidth}
 \includegraphics[width=1\textwidth]{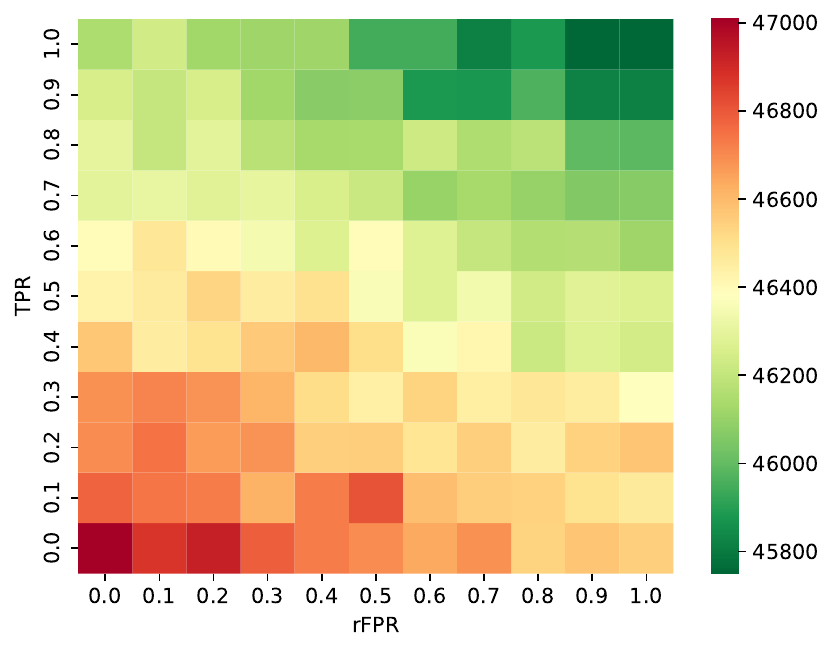}
 \caption{Rerostering cost}
 \label{img:multi-re}
\end{subfigure}
\begin{subfigure}{0.44\textwidth}
 \includegraphics[width=1\textwidth]{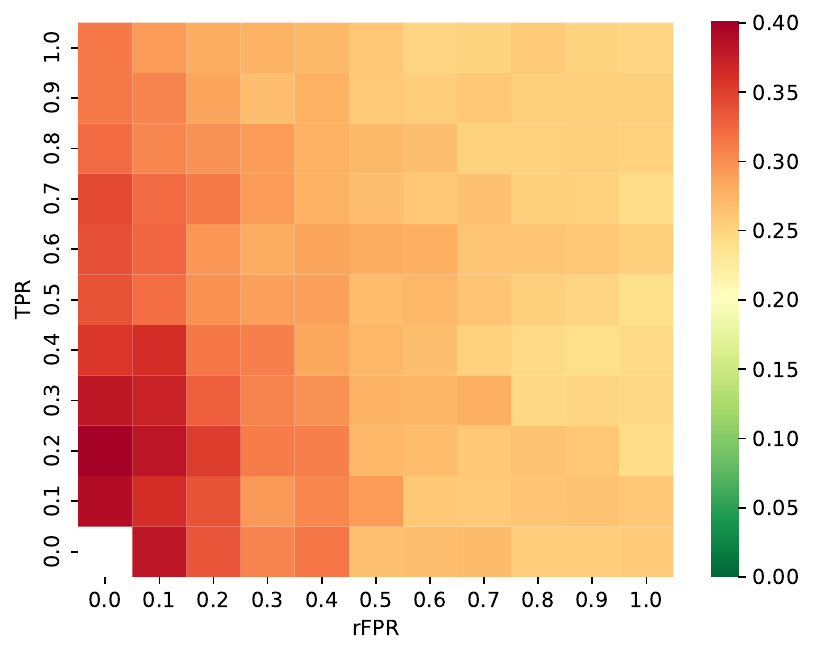}
 \caption{\% of reserve shifts changed}
 \label{img:multi-res}
\end{subfigure}
\begin{subfigure}{0.44\textwidth}
 \includegraphics[width=1\textwidth]{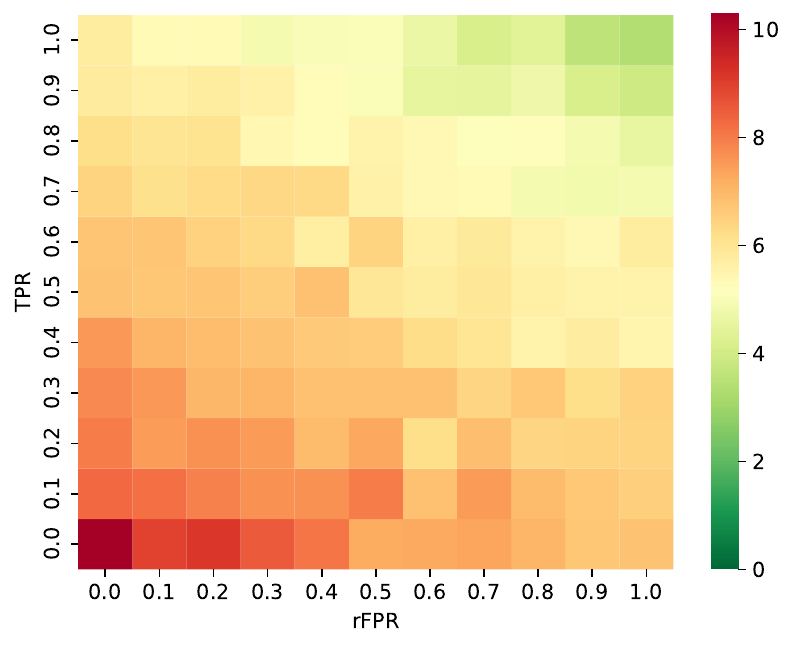}
 \caption{Number of working shifts changed}
 \label{img:multi-work}
\end{subfigure}
\begin{subfigure}{0.44\textwidth}
 \includegraphics[width=1\textwidth]{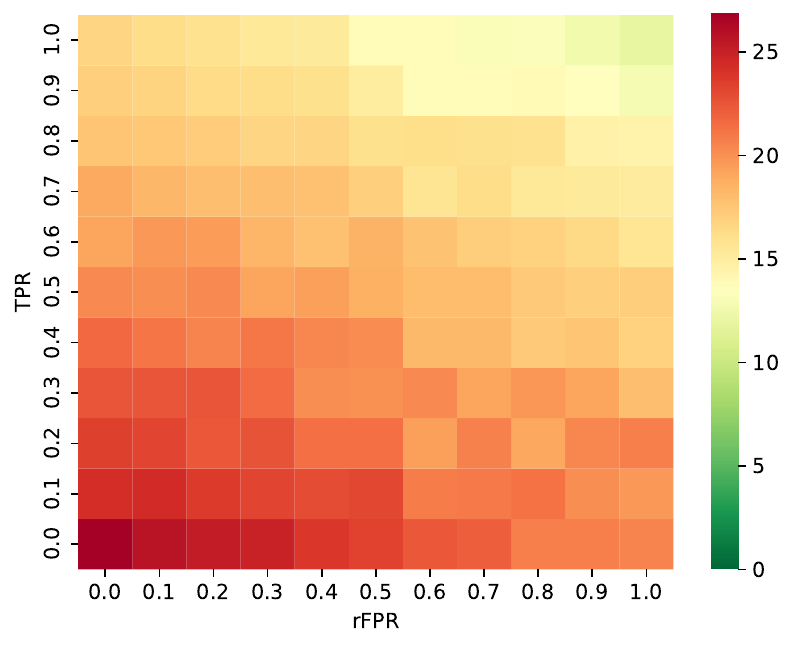}
 \caption{Number of days off changed}
 \label{img:multi-off}
\end{subfigure}
\caption{Number of scheduled reserve shifts and the number of changes made when rerostering for various prediction model performance levels for the problem instance with hierarchical skills.}
\label{img:multi-changes}
\end{figure}

Figure \ref{img:multi-re} shows how the rerostering cost increases as the TPR and rFPR decrease.
Without accurate predictions concerning which nurses will be absent, and thus which skills will need to be covered, reserve shifts become less effective during rerostering. 
Nevertheless, it is possible to observe the same phenomenon observed in the setting with uniform skills: in addition to the TPR, a higher rFPR also contributes to an overall reduction of the rerostering cost. 
Due to the cost structure considered, an excessive number of nurses in reserve shifts makes it less costly to repair the roster without disrupting other nurses' schedules.

Figures \ref{img:multi-res}, \ref{img:multi-work} and \ref{img:multi-off} show the number of changes to reserve shifts, working shifts and days off for different values of the TPR and rFPR.
In general, fewer reserve shifts can be converted compared to the scenario with uniform skills.
The roster contains a comparable number of reserve shifts, but due to the additional restrictions imposed by skills, the rerostering method makes less effective use of the available reserve shifts.
This is also reflected in the higher number of changes to working shifts and days off compared to the scenario with uniform skills.
In general, it is much more important to accurately predict precisely which nurses will become absent when considering hierarchical skills compared to the scenario with uniform skills, where it was almost always beneficial to schedule more reserve shifts. 

\begin{figure}[hptb]
\centering
\begin{subfigure}{0.45\textwidth}
 \includegraphics[width=1\textwidth]{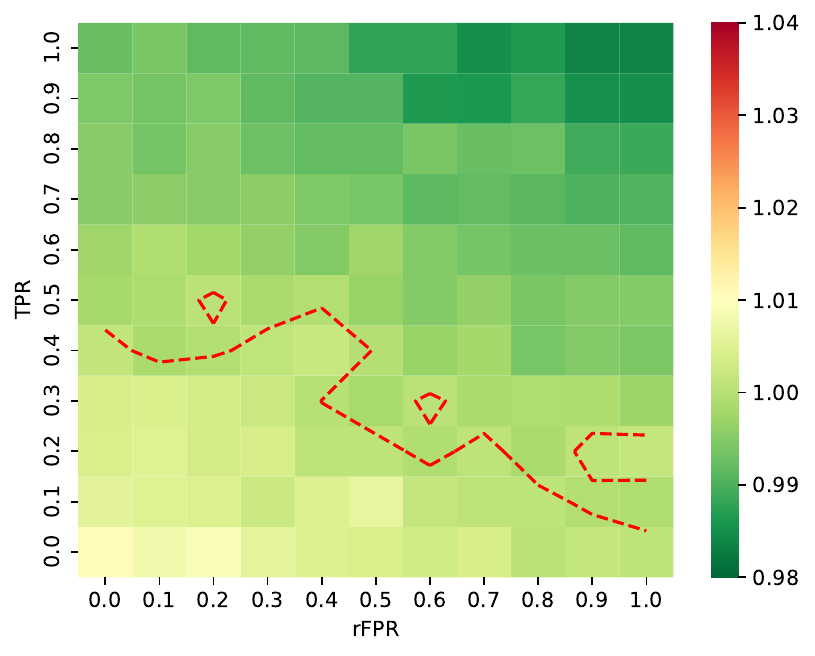}
 \caption{One reserve shift per day}
\end{subfigure}
\begin{subfigure}{0.45\textwidth}
 \includegraphics[width=1\textwidth]{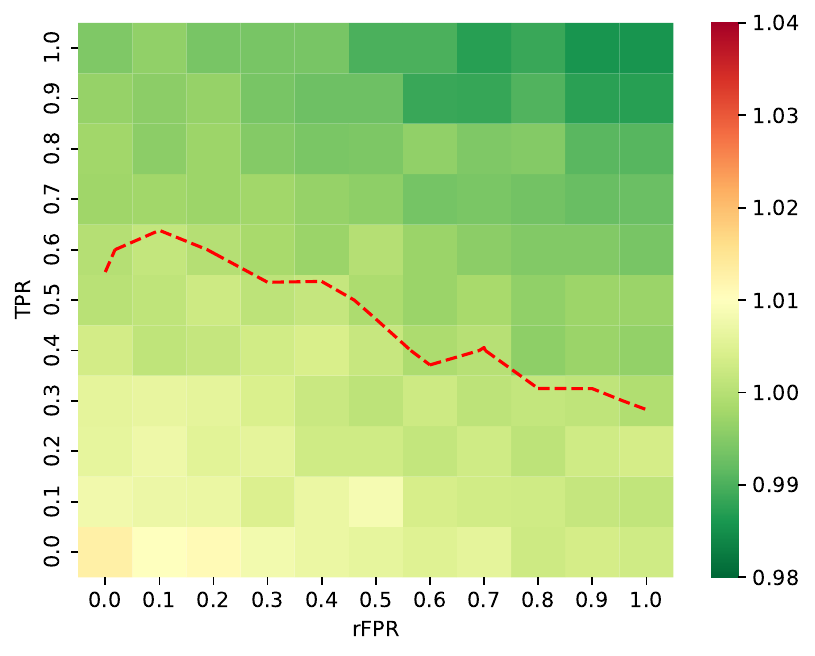}
 \caption{Two reserve shifts per day}
\end{subfigure}
\begin{subfigure}{0.45\textwidth}
 \includegraphics[width=1\textwidth]{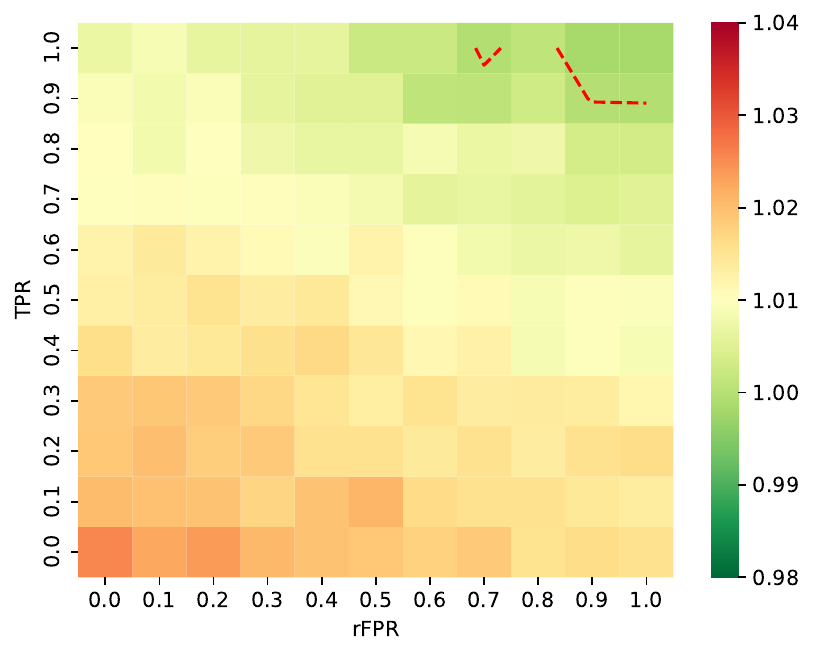}
 \caption{Three reserve shifts per day}
\end{subfigure}
\begin{subfigure}{0.45\textwidth}
 \includegraphics[width=1\textwidth]{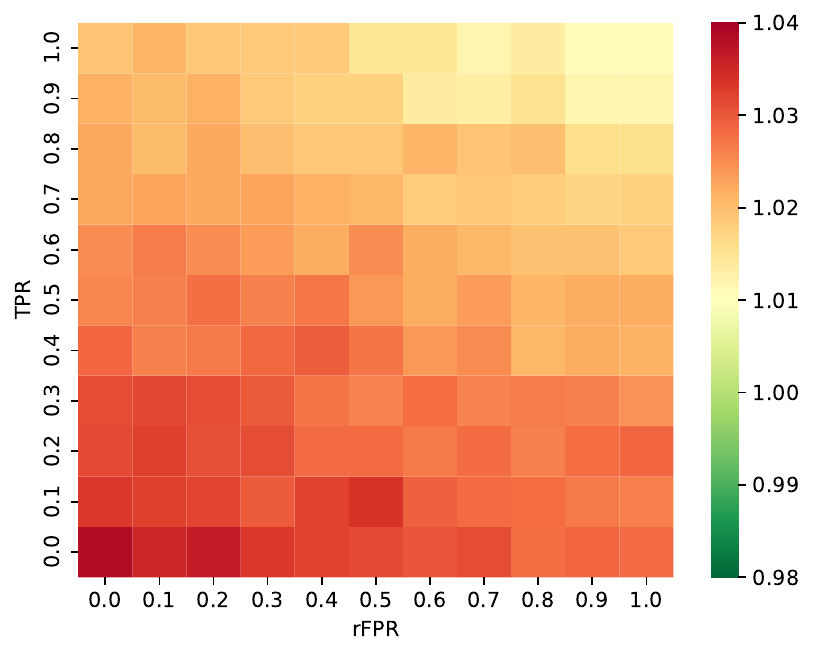}
 \caption{Four reserve shifts per day}
\end{subfigure}
\caption{Rerostering cost of the ML-informed approach over the rerostering cost of the baseline policies for the problem instance with hierarchical skills.}
\label{img:multi-comp}
\end{figure}

Figure \ref{img:multi-comp} compares the performance of the ML-informed robust rostering approach to the four baseline policies.
The reported value is the ratio of the rostering cost obtained by the ML-informed approach over the rerostering cost obtained by the baseline policy.
The red dashed line denotes when the ratio is equal to one: the level at which the rerostering cost of both methods is the same.
Better results are obtained by the ML-informed approach compared to the policy that assigns one nurse per day to a reserve shift for relatively low TPR and rFPR levels.
However, as the fixed number of reserve shifts per day increases, it becomes increasingly difficult for the ML-informed approach to generate comparable solutions, to the point that the ML-informed approach never manages to outperform the four-shift policy, even when it is able to make perfect predictions concerning the absences.
This result indicates that, when considering hierarchical skills, the ML model should not only predict the number of reserve shifts per day, but also to which employee types these reserve shifts must be assigned.

\section{Conclusions}
\label{s:conc}

Reserve shift buffers are commonly used to increase the robustness of a roster.
While assigning reserve shifts to employees is typically less costly than overtime or employing external employees, the number of reserve shifts required on each day must still be carefully determined.
Rather than relying on a human expert or extensive computer simulations, we consider the use of a predict-then-optimize approach to determine an appropriate number.
By predicting employee absences, we derive a suitable number of reserve shifts that must be scheduled on each day.

The core contribution of this paper is methodological in nature and centers around computing the robustness of a roster generated by the predict-then-optimize approach, assuming the ML model can make predictions at a predetermined performance level.
The ML model is characterized by its True Positive Rate and False Positive Rate.
By carefully interpreting these performance metrics, we were able to simulate the model's predictions concerning employee absenteeism.
The key advantage of simulating predictions, instead of actually training and testing ML models, is that we do not rely on the availability of data concerning the employees.

Building upon this new methodology, we proposed an approach to determine minimum performance requirements necessary to obtain rosters that are more robust than those generated by simple non-data-driven policies.
We evaluate the predict-then-optimize approach on a well-known nurse rostering problem data set.
When all nurses have identical skills, and thus exhibit a large degree of substitutability, the predict-then-optimize approach outperforms the non-data-driven policies with reasonably low performance requirements.
The results demonstrate how ML models with a high False Positive Rate can compensate for a low True Positive Rate due to the flexibility the reserve shifts induce during rerostering.
For problem instances with a hierarchical skill structure, the minimum performance requirements increase.
Our results demonstrate how predicting absences of individual employees results in more robust rosters compared to simply predicting the total number of absent nurses.
However, to do so may require additional data for training and testing a suitable ML model.

The emphasis in our work is on the use of reserve shift buffers to increase roster robustness.
The computational experiments resulted in new insights concerning where best to include reserve shifts in the rosters.
Future research may build upon these insights to define new non-data-driven robust rostering policies that do not require dedicated ML models or sophisticated optimization models to generate robust rosters.
Finally, we focused on how predictions of a binary classifier can be simulated.
By generalizing our methodology and applying it to other prediction tasks, it may be possible to derive similar minimum performance requirements for predict-then-optimize approaches to other assignment, sequencing or scheduling problems.

\subsection*{Acknowledgments}
This research was partially supported by KU Leuven C24E/23/012 ``Human-centred decision support based on new theory for personnel rostering''.
Editorial consultation provided by Luke Connolly (KU Leuven).

\bibliographystyle{abbrvnat}
\bibliography{bibliography.bib}

\begin{thebibliography}{21}
\providecommand{\natexlab}[1]{#1}
\providecommand{\url}[1]{\texttt{#1}}
\expandafter\ifx\csname urlstyle\endcsname\relax
  \providecommand{\doi}[1]{doi: #1}\else
  \providecommand{\doi}{doi: \begingroup \urlstyle{rm}\Url}\fi

\bibitem[Bamberg et~al.(2012)Bamberg, Dettmers, Funck, Kr{\"a}he, and
  Vahle-Hinz]{bamberg2012effects}
E.~Bamberg, J.~Dettmers, H.~Funck, B.~Kr{\"a}he, and T.~Vahle-Hinz.
\newblock Effects of on-call work on well-being: Results of a daily survey.
\newblock \emph{Applied Psychology: Health and Well-Being}, 4\penalty0
  (3):\penalty0 299--320, 2012.

\bibitem[Becker et~al.(2019)Becker, Steenweg, and Werners]{becker2019cyclic}
T.~Becker, P.~M. Steenweg, and B.~Werners.
\newblock Cyclic shift scheduling with on-call duties for emergency medical
  services.
\newblock \emph{Health care management science}, 22:\penalty0 676--690, 2019.

\bibitem[Ceschia et~al.(2019)Ceschia, Dang, De~Causmaecker, Haspeslagh, and
  Schaerf]{ceschia2019}
S.~Ceschia, N.~Dang, P.~De~Causmaecker, S.~Haspeslagh, and A.~Schaerf.
\newblock The second international nurse rostering competition.
\newblock \emph{Annals of Operations Research}, 274\penalty0 (1-2):\penalty0
  171--186, 2019.

\bibitem[Dillon and Kontogiorgis(1999)]{dillon1999us}
J.~E. Dillon and S.~Kontogiorgis.
\newblock {US Airways} optimizes the scheduling of reserve flight crews.
\newblock \emph{Interfaces}, 29\penalty0 (5):\penalty0 123--131, 1999.

\bibitem[El-Rifai et~al.(2016)El-Rifai, Garaix, and Xie]{el2016proactive}
O.~El-Rifai, T.~Garaix, and X.~Xie.
\newblock Proactive on-call scheduling during a seasonal epidemic.
\newblock \emph{Operations Research for Health Care}, 8:\penalty0 53--61, 2016.

\bibitem[Elmachtoub and Grigas(2022)]{elmachtoub2022}
A.~N. Elmachtoub and P.~Grigas.
\newblock Smart “predict, then optimize”.
\newblock \emph{Management Science}, 68\penalty0 (1):\penalty0 9--26, 2022.

\bibitem[Farrington et~al.(2023)Farrington, Li, Wong, and
  Utley]{farrington2023}
J.~Farrington, K.~Li, W.~K. Wong, and M.~Utley.
\newblock Reducing platelet wastage with a machine learning-based allocation
  policy.
\newblock In \emph{Proceedings of the 49th Annual Meeting of the {EURO} Working
  Group on Operational Research Applied to Health Services (ORAHS)}, Graz,
  Austria, July 2023.

\bibitem[Grinza and Rycx(2020)]{grinza2020impact}
E.~Grinza and F.~Rycx.
\newblock {The impact of sickness absenteeism on firm productivity: New
  evidence from Belgian matched employer--employee panel data}.
\newblock \emph{Industrial Relations: A Journal of Economy and Society},
  59\penalty0 (1):\penalty0 150--194, 2020.

\bibitem[Hudson and Shen(2015)]{hudson2015}
C.~K. Hudson and W.~Shen.
\newblock Understaffing: An under-researched phenomenon.
\newblock \emph{Organizational Psychology Review}, 5\penalty0 (3):\penalty0
  244--263, 2015.

\bibitem[Ingels and Maenhout(2015)]{ingels2015}
J.~Ingels and B.~Maenhout.
\newblock The impact of reserve duties on the robustness of a personnel shift
  roster: An empirical investigation.
\newblock \emph{Computers \& Operations Research}, 61:\penalty0 153--169, 2015.
\newblock ISSN 0305-0548.

\bibitem[Ingels and Maenhout(2018)]{ingels2018}
J.~Ingels and B.~Maenhout.
\newblock The impact of overtime as a time-based proactive scheduling and
  reactive allocation strategy on the robustness of a personnel shift roster.
\newblock \emph{Journal of Scheduling}, 21:\penalty0 143--165, 2018.

\bibitem[Ingels and Maenhout(2019)]{ingels2019}
J.~Ingels and B.~Maenhout.
\newblock Optimised buffer allocation to construct stable personnel shift
  rosters.
\newblock \emph{Omega}, 82:\penalty0 102--117, 2019.

\bibitem[Kocakulah et~al.(2016)Kocakulah, Kelley, Mitchell, Ruggieri,
  et~al.]{kocakulah2016}
M.~C. Kocakulah, A.~G. Kelley, K.~M. Mitchell, M.~P. Ruggieri, et~al.
\newblock Absenteeism problems and costs: causes, effects and cures.
\newblock \emph{International Business \& Economics Research Journal},
  15\penalty0 (3):\penalty0 89--96, 2016.

\bibitem[Kuhn(2022)]{kuhn2022}
M.~Kuhn.
\newblock \emph{caret: Classification and Regression Training}, 2022.
\newblock URL \url{https://CRAN.R-project.org/package=caret}.
\newblock R package version 6.0-93.

\bibitem[Potthoff et~al.(2010)Potthoff, Huisman, and
  Desaulniers]{potthoff2010column}
D.~Potthoff, D.~Huisman, and G.~Desaulniers.
\newblock Column generation with dynamic duty selection for railway crew
  rescheduling.
\newblock \emph{Transportation Science}, 44\penalty0 (4):\penalty0 493--505,
  2010.

\bibitem[{SD Worx}(2022)]{SDWorx2022}
{SD Worx}.
\newblock {Absenteeism rises to new record in 2022}.
\newblock
  \url{https://www.sdworx.be/nl-be/over-sd-worx/pers/2023-01-17-ziekteverzuim-stijgt-naar-nieuw-record-2022},
  2022.
\newblock Accessed on December 12, 2023.

\bibitem[Tarro et~al.(2020)Tarro, Llaurad{\'o}, Ulldemolins, Hermoso, and
  Sol{\`a}]{tarro2020}
L.~Tarro, E.~Llaurad{\'o}, G.~Ulldemolins, P.~Hermoso, and R.~Sol{\`a}.
\newblock Effectiveness of workplace interventions for improving absenteeism,
  productivity, and work ability of employees: a systematic review and
  meta-analysis of randomized controlled trials.
\newblock \emph{International journal of environmental research and public
  health}, 17\penalty0 (6):\penalty0 1901, 2020.

\bibitem[Ticharwa et~al.(2019)Ticharwa, Cope, and Murray]{ticharwa2019}
M.~Ticharwa, V.~Cope, and M.~Murray.
\newblock Nurse absenteeism: An analysis of trends and perceptions of nurse
  unit managers.
\newblock \emph{Journal of Nursing Management}, 27\penalty0 (1):\penalty0
  109--116, 2019.

\bibitem[Tulabandhula and Rudin(2013)]{tulabandhula2013}
T.~Tulabandhula and C.~Rudin.
\newblock Machine learning with operational costs.
\newblock \emph{Journal of Machine Learning Research}, 14:\penalty0 1989--2028,
  2013.

\bibitem[Wickert et~al.(2019)Wickert, Smet, and Vanden~Berghe]{wickert2019}
T.~I. Wickert, P.~Smet, and G.~Vanden~Berghe.
\newblock The nurse rerostering problem: Strategies for reconstructing
  disrupted schedules.
\newblock \emph{Computers \& Operations Research}, 104:\penalty0 319--337,
  2019.

\bibitem[Wickert et~al.(2021)Wickert, Smet, and Vanden~Berghe]{wickert2021}
T.~I. Wickert, P.~Smet, and G.~Vanden~Berghe.
\newblock Quantifying and enforcing robustness in staff rostering.
\newblock \emph{Journal of Scheduling}, 24\penalty0 (3):\penalty0 347--366,
  2021.

\end{thebibliography}

\clearpage

\appendix

\setcounter{table}{0}
\renewcommand{\thetable}{A.\arabic{table}}
\setcounter{figure}{0}
\renewcommand{\thefigure}{A.\arabic{figure}}

\section{Rostering problem MIP formulation}
\label{ss:ro-mip}

Table \ref{tab:rostering} provides the notation used in the MIP formulation of the robust rostering problem.
For each employee $n \in N$, day $d \in D$, shift $s \in S$ and employee skill $k \in K_n$, let $x_{ndsk}$ be a binary variable which equals 1 if employee $n$ is assigned to shift $s$ on day $d$ in skill $k$, and $0$ otherwise.
Overtime of employee $n \in N$ is penalized using variable $v_n^5 \in \mathbb{N}_{\geq 0}$, which equals the number of days worked over the maximum allowed for employee $n$.
Violations of the minimum demand requirement for shift $s \in S_w$ and skill $k \in K$ on day $d \in D$ are penalized using variable $v_{dsk}^6 \in \mathbb{N}_{\geq 0}$, which equals the number of employees below the minimum required for day $d$, shift $s$ and skill $k$.
The MIP formulation of the rostering problem is given by Equations \eqref{eq:OF_ro}-\eqref{eq:ro_bounds3}.

\begin{table}[hptb]
 \footnotesize
 \centering
 \begin{tabular}{p{0.1\textwidth}p{0.9\textwidth}}
 \toprule
 \multicolumn{2}{l}{\textbf{Sets}} \\
 \midrule
 $N$ & Set of employees, indexed by $n$ \\
 $D$ & Set of days in the scheduling period, indexed by $d$ \\
 $H $ & Set of days in the preceding scheduling period, indexed by $h$ \\
 $S$ & Set of shifts, indexed by $s$ \\
 $S_w \subseteq S$ & Subset of working shifts in $S$, excluding the reserve shift\\
 $\Tilde{S}$ & Set of forbidden shift successions $(s_1, s_2)$ \\
 $K$ & Set of employee skills, indexed by $k$ \\
 $K_n \subset K$ & Subset of skills for which employee $n$ is qualified\\
 $N_k \subseteq N$ & Subset of employees who are qualified for skill $k$ \\
 $U$ & Set of tuples $(n,d,s)$ that define forbidden assignments of employee $n$ to shift $s$ on day $d$ \\
 \midrule
 \multicolumn{2}{l}{\textbf{Parameters}} \\
 \midrule
 $s_n$ & Index of the night shift in $S$ \\
 $m_{dsk}$ & Minimum number of employees required on day $d$ for shift $s$ with skill $k$ \\
 $\beta^1_n$ & Maximum number of consecutive working days for employee $n$ \\
 $\beta^2_n$ & Maximum number of consecutive night shifts for employee $n$ \\
 $\beta^3_n$ & Minimum number of working days in the scheduling period for employee $n$ \\
 $\beta^5_n$ & Maximum number of working days in the scheduling period for employee $n$ \\
 $\beta^6_n$ & Maximum number of reserve shifts in the scheduling period for employee $n$ \\
 $\hat{x}_{nhs}$ & Binary parameter that equals 1 is employee $n$ was assigned to shift $s$ on day $h$ in the preceding scheduling period \\
 \midrule
 \multicolumn{2}{l}{\textbf{Costs and penalties}} \\
 \midrule
 $\omega^1_n$ & Wage cost for assigning any working shift to employee $n$ \\
 $\omega^5_n$ & Overtime wage cost for employee $n$ for any additional shift worked over their maximum \\
 $\omega^6$ & Cost of understaffing a shift \\
 \bottomrule
 \end{tabular}
 \caption{Notation used in the robust rostering MIP formulation.}
 \label{tab:rostering}
\end{table}

{\allowdisplaybreaks
\begin{align}
\min & \rlap{$\displaystyle\sum_{n \in N}\sum_{d \in D}\sum_{s \in S_w}\sum_{k \in K_n} x_{ndsk}\omega^1_n + \sum_{n \in N} v^5_n\omega^5_n + \sum_{d \in D}\sum_{s \in S_w}\sum_{k \in K}v^6_{dsk}\omega^6 $} \label{eq:OF_ro} \\
s.t.\ & \sum_{s \in S}\sum_{k \in K_n}x_{ndsk} \leq 1 & \forall n \in N, d \in D \label{eq:unicity_shift} \\
& \sum_{n \in N_k}x_{ndsk} + v^6_{dsk} \geq m_{dsk} & \forall d \in D, s \in S_w, k \in K \label{eq:minimum_employees} \\
& \sum_{k \in K_n}(x_{nds_1k} + x_{n(d+1)s_2k}) \leq 1 & \forall n \in N, d \in D\setminus \left\{|D|\right\}, \nonumber \\[-15pt] 
& & (s_1,s_2) \in \tilde{S} \label{eq:forbidden_shifts_successions} \\
& \sum_{k \in K_n}(\hat{x}_{n(-1)s_1} + x_{n0s_2k}) \leq 1 & \forall n \in N, (s_1,s_2) \in \tilde{S} \label{eq:forbidden_shifts_successions_hist} \\
& \sum_{d' = d}^{\beta_n^1 + d}\sum_{s \in S}\sum_{k \in K_n}x_{nd'sk} \leq \beta^1_n & \forall n \in N, d \in {1, ..., |D|-\beta^1_n} \label{eq:max_consecutive_working_days} \\
& \sum_{h = \Delta - \beta_n^1}^{0}\sum_{s \in S}\hat{x}_{nh s} + \sum_{d=0}^{\Delta}\sum_{s \in S}\sum_{k \in K_n}x_{ndsk}\leq \beta^1_n & \forall n \in N, \Delta \in \left\{0, \beta^1_n\right\} \label{eq:max_consecutive_working_days_hist} \\ 
& \sum_{d' = d}^{\beta_n^2 + d}\sum_{k \in K_n}x_{nd's_nk} \leq \beta^2_n & \forall n \in N, d \in {1, ..., |D|-\beta^2_n} \label{eq:max_consecutive_nights} \\
& \sum_{h = \Delta - \beta_n^2}^{0}\hat{x}_{nh s_n} + \sum_{d=0}^{\Delta}\sum_{k \in K_n}x_{nds_nk} \leq \beta^2_n & \forall n \in N, \Delta \in \left\{0, \beta^2_n\right\} \label{eq:max_consecutive_nights_hist} \\ 
& \sum_{k \in K_n}x_{ndsk} = 0 & \forall (n,d,s) \in U \label{eq:undesired_shifts}\\
& \sum_{d \in D}\sum_{s \in S_w}\sum_{k \in K_n}x_{ndsk} \geq \beta^3_n & \forall n \in N \label{eq:min_working_days} \\
& \sum_{d \in D}\sum_{s \in S_w}\sum_{k \in K_n}x_{ndsk} - v^5_n\leq \beta^5_n & \forall n \in N \label{eq:max_working_days} \\
& x_{ndsk} \in \{0, 1\} & \forall n \in N, d \in D, s \in S, k \in K_n \label{eq:ro_bounds1} \\
& v^5_n \geq 0 & \forall n \in N \label{eq:ro_bounds2} \\
& v^6_{dsk} \geq 0 & \forall d \in D, s \in S, k \in K_n \label{eq:ro_bounds3}
\end{align}
}

Objective function \eqref{eq:OF_ro} minimizes a weighted sum of three components: (i) employees' regular wages, (ii) overtime costs and (iii) understaffing costs. 
Constraints \eqref{eq:unicity_shift} ensure that each employee is assigned to at most one shift per day. 
Constraints \eqref{eq:minimum_employees} ensure the minimum required number of employees with certain skills on each day and shift as a soft constraint whose violation is penalized by the $v^6_{dks}$ variables.
Constraints \eqref{eq:forbidden_shifts_successions} and \eqref{eq:forbidden_shifts_successions_hist} ensure that no forbidden shift successions occur, taking into account the end of the previous scheduling period. 
Similarly, Constraints \eqref{eq:max_consecutive_working_days} and \eqref{eq:max_consecutive_working_days_hist} limit the maximum number of consecutive working days. 
Constraints \eqref{eq:max_consecutive_nights} and \eqref{eq:max_consecutive_nights_hist} ensure the maximum number of consecutive night shifts is never exceeded. 
Constraints \eqref{eq:undesired_shifts} prevent the assignment of shifts in which employees cannot work. 
Constraints \eqref{eq:min_working_days} and \eqref{eq:max_working_days} limit the minimum and the maximum number of assignments in the scheduling period for each employee.
Finally, Constraints \eqref{eq:ro_bounds1}-\eqref{eq:ro_bounds3} define bounds on the decision variables.

\section{Rerostering problem MIP formulation}
\label{ss:re-mip}
\setcounter{table}{0}
\renewcommand{\thetable}{B.\arabic{table}}
\setcounter{figure}{0}
\renewcommand{\thefigure}{B.\arabic{figure}}

The rerostering problem takes into account the realization of the absences per day and forcibly prevents the assignment of absent employees to any shift. 
The same contractual constraints as in the rostering problem must be respected.
The most preferable way of repairing a roster is by transforming a reserve shift into an actual shift. 
However, if this is not possible, then the roster can also be repaired by changing other working shifts.
The least preferable option is to call in personnel that have the day off. 
While utilizing reserve shifts is always preferred, these other two methods of repairing a roster can be used if a feasible solution is otherwise unattainable.
If an absent employee was originally assigned to a reserve shift, and that reserve shift has not yet been converted into a working shift, their assignment to the reserve shift is maintained.
The rerostering objective function includes the same cost minimization terms from the robust rostering model, in addition to a weighted term that minimizes the number of changes with respect to the original roster.

Table \ref{tab:rerostering} provides an overview of the notation used in the rerostering MIP formulation.
Three decision variables count the number of changes made to the original roster.
For each employee $n \in N$ and day $d \in D$, $v_{nd}^3 \in \mathbb{N}_{\geq 0}$ counts the number of shift changes compared to the original roster excluding the reserve shift for employee $n$ on day $d$.
Similarly, $v_{nd}^2 \in \mathbb{N}_{\geq 0}$ counts the number of reserve shifts that have been converted into working shifts, while $v_{nd}^4 \in \mathbb{N}_{\geq 0}$ counts the number of times a day off is replaced with a working shift or vice versa.
For each employee $n \in N$, day $d \in D$ and shift $s \in S$, binary variable $y^{'}_{nds}$ equals 1 if employee $n$ is assigned to shift $s$ on day $d$ in either the original or new roster, and $0$ otherwise.
Two auxiliary variables $y^{''}_{nds}$ and $y^{'''}_{nd}$ keep track of the number of changes compared to the original roster, for a given nurse $n$ day $d$ and shift $s$ combination, and for a given nurse $n$ and day $d$ pair, respectively.
The MIP formulation of the rerostering problem is given by Equations \eqref{eq:OF_re}-\eqref{eq:re_bounds4}.

\begin{table}[hptb]
 \footnotesize
 \centering
 \begin{tabular}{p{0.13\textwidth}p{0.81\textwidth}}
 \toprule
 \multicolumn{2}{l}{\textbf{Sets}} \\
 \midrule
 $\hat{N}$ & Set of absent employees \\
 \midrule
 \multicolumn{2}{l}{\textbf{Parameters}} \\
 \midrule
 $\hat{c}_{nd} \in \left\{ 0, 1\right\}$ & Binary parameter equal to $1$ if employee $n$ is absent on day $d$, $0$ otherwise \\
 $c_{ndsk} \in \left\{ 0, 1\right\}$ & Binary parameter equal to $1$ if employee $n$ has been assigned to shift $s$ on day $d$ with skill $k$ in the original roster, $0$ otherwise \\
 \midrule
 \multicolumn{2}{l}{\textbf{Costs and penalties}} \\
 \midrule
 $\omega^2_n$ & Cost incurred when converting the shift assigned to employee $n$ into another working shift \\
 $\omega^3_n$ & Cost incurred when converting the reserve shift assigned to employee $n$ into a working shift\\
 $\omega^4_n$ & Cost incurred when converting the working shift assigned to employee $n$ into a day off, or vice versa \\
 \bottomrule
 \end{tabular}
 \caption{Sets and parameters used in the rerostering MIP formulation.}
 \label{tab:rerostering}
\end{table}

{\allowdisplaybreaks
\begin{align}
\min\ & \eqref{eq:obj_complete} + \sum_{n \in N}\sum_{d \in D}\sum_{i \in \left\{2, 3, 4\right\}}v^i_{nd}\omega^i_n & \label{eq:OF_re} \\
s.t.\ & \hat{c}_{nd} + \sum_{s \in S}\sum_{k \in K_n}x_{ndsk} \leq 1 & \forall n \in N, d \in D \label{eq:not_assignment_of_absentees}\\
& \sum_{k \in K_n}(c_{ndsk} + x_{ndsk}) \leq 2y^{'}_{nds} & \forall n \in N\setminus \hat{N}, d \in D, \nonumber \\[-15pt]
& & s \in S \label{eq:changes_calculation1}\\
& \sum_{k \in K_n}(c_{ndsk} + x_{ndsk}) + y^{''}_{nds} \geq 2y^{'}_{nds} & \forall n \in N\setminus \hat{N}, d \in D, \nonumber \\[-15pt] 
& & s \in S \label{eq:changes_calculation2}\\
& \sum_{s \in S}y^{''}_{nds} - 2y^{'''}_{nd} \leq 0 & \forall n \in N\setminus \hat{N}, d \in D \label{eq:changes_calculation3}\\
& \sum_{s \in S_w}\sum_{k \in K_n}(c_{ndsk} + x_{ndsk}) -1 - v^2_{nd} \leq 1 - y^{'''}_{nd} & \forall n \in N\setminus\hat{N}, d \in D \label{eq:shift_changes_not_reserve}\\ 
& \sum_{s \in S}\sum_{k \in K_n}x_{ndsk} + \sum_{k \in K_n}c_{nds'k} -1 - v^3_{nd} \leq 1 - y^{'''}_{nd} & \forall n \in N\setminus\hat{N}, d \in D \label{eq:shift_changes_reserve}\\ 
& \sum_{s \in S}\sum_{k \in K_n}c_{ndsk} + \sum_{s^{''} \in S_w}\sum_{k \in K_n} x_{nds^{''}k} + \nonumber \\
& \qquad\qquad\qquad\qquad v^4_{nd} \geq 2(y^{'''}_{nd} - \sum_{k \in K_n}c_{nds'k}) & \forall n \in N\setminus \hat{N}, d \in D \label{eq:day_off_changes}\\
& \sum_{s \in S}\sum_{k \in K_n}x_{ndsk} \geq \sum_{k \in K_n}c_{nds'k} & \forall n \in N\setminus\hat{N}, d \in D \label{eq:block_reserve_to_day_off} \\
& x_{ndsk} \in \{0, 1\} & \forall n \in N, d \in D, s \in S, k \in K_n \label{eq:re_bounds1} \\
& v^2_{nd}, v^3_{nd}, v^4_{nd} \geq 0 & \forall n \in N, d \in D \label{eq:re_bounds2} \\
& y^{\prime}_{nds}, y^{\prime\prime}_{nds} \in \{0, 1\} & \forall n \in N, d \in D, s \in S \label{eq:re_bounds3} \\
& y^{\prime\prime\prime}_{nd} \in \{0, 1\} & \forall n \in N, d \in D \label{eq:re_bounds4}
\end{align} 
}

Objective function \eqref{eq:OF_re} adds three additional components to the objective function \eqref{eq:OF_ro} of the robust rostering problem: the cost of changes made to the roster with respect to the original schedule (before the realization of the absent shifts), the cost of converting a reserve shift into a working shift and the cost of converting a day off into a working shift.
Constraints \eqref{eq:not_assignment_of_absentees} prevent employees that are absent on day $d$ to be assigned to a working shift on that day. 
Constraints \eqref{eq:changes_calculation1}-\eqref{eq:changes_calculation3} calculate the number of changes compared to the original roster, storing them in auxiliary variables. 
Constraints \eqref{eq:shift_changes_not_reserve} calculate working shift changes, while Constraints \eqref{eq:shift_changes_reserve} compute the number of reserve shifts converted into working shifts. 
Constraints \eqref{eq:day_off_changes} count how many times a day off is converted into a working day (or vice versa). 
Constraints \eqref{eq:block_reserve_to_day_off} prevent transforming any reserve shift into a day off.
Finally, Constraints \eqref{eq:re_bounds1}-\eqref{eq:re_bounds4} define bounds on the decision variables.

\end{document}